\documentclass[letterpaper, 10 pt, conference]{ieeeconf}  
\usepackage{amsmath}
\usepackage{graphicx}
\usepackage{rotating}
\usepackage{color}

\IEEEoverridecommandlockouts                              

\usepackage{graphics} 
\usepackage{epsfig} 
\usepackage{mathptmx} 
\usepackage{times} 
\usepackage{amsmath} 
\usepackage{amssymb}  
\usepackage{subfig}

\newcommand*{\permcomb}[4][0mu]{{{}^{#3}\mkern#1#2_{#4}}}

\newcommand*{\comb}[1][-1mu]{\permcomb[#1]{C}}
\newcommand*{\etal}{\textit{et~al.}}

\newtheorem{theorem}{\bf Theorem}[section]

\newcommand{\qed}{\nobreak \ifvmode \relax \else
      \ifdim\lastskip<1.5em \hskip-\lastskip
      \hskip1.5em plus0em minus0.5em \fi \nobreak
      \vrule height0.75em width0.5em depth0.25em\fi}

\def\baselinestretch{0.999}

\title{\LARGE \bf
Dynamic Body VSLAM with Semantic Constraints
}

\author{N. Dinesh Reddy $^{1}$, Prateek Singhal$^{2}$, Visesh Chari$^{1,3}$ and  K. Madhava Krishna$^{1}$
\thanks{$^{1}$ with affiliation to International Institute of Technology, Hyderabad}
\thanks{ $^{2}$ with affiliation to Georgia Institute of Technology, Atlanta}
\thanks{$^{3}$ with affiliation to the WILLOW group, INRIA, Paris}
}

\begin{document}

\maketitle
\thispagestyle{empty}
\pagestyle{empty}

\begin{abstract}
  Image based reconstruction of urban environments is a challenging problem that deals with
  optimization of large number of variables, and has several sources of errors like the presence of 
  dynamic objects.  Since most large scale approaches 
  make the assumption of observing static scenes, dynamic objects are relegated to the noise
  modeling section of such systems. This is an approach of convenience since the RANSAC based framework used to compute most
  multiview geometric quantities for static scenes naturally confine dynamic objects to the class of outlier
  measurements. However, reconstructing dynamic objects along with the static environment helps us
  get a complete picture of an urban environment. Such understanding can then be used for important
  robotic tasks like path planning for autonomous navigation, obstacle tracking and avoidance,
  and other areas.

  In this paper, we propose a system for robust SLAM that works in both static and dynamic
  environments. To overcome the challenge of dynamic objects in the scene, we propose a new model
  to incorporate \emph{semantic constraints} into the reconstruction algorithm. While some of these
  constraints are based on multi-layered dense CRFs trained over appearance \emph{as well as motion
  cues}, other proposed constraints can be expressed as additional terms in the bundle adjustment
  optimization process that does iterative refinement of 3D structure and camera / object motion
  trajectories. We show results on the challenging KITTI urban dataset for accuracy of motion
  segmentation and reconstruction of the trajectory and shape of moving objects relative to ground
  truth. We are able to show average relative error reduction by a significant amount for moving
  object trajectory reconstruction relative to state-of-the-art methods like VISO 2\cite{lenz},
  as well as standard bundle adjustment algorithms.

\end{abstract}


\section{\bf INTRODUCTION}

Vision based SLAM (vSLAM) is becoming an increasingly widely researched problem, partly because of its
ability to produce good quality reconstructions with affordable hardware, and partly
because of increasing computational power that results in computational affordability of
huge optimization problems. While vSLAM systems are maturing and getting progressively
complicated, the two main components remain camera localization (or camera pose estimation) and 3D reconstruction.
Generally, these two components precede an optimization based joint refinement of both
camera pose and 3D structure, called bundle adjustment.

\begin{figure}
\centering
\includegraphics[width=85mm,height=45mm]{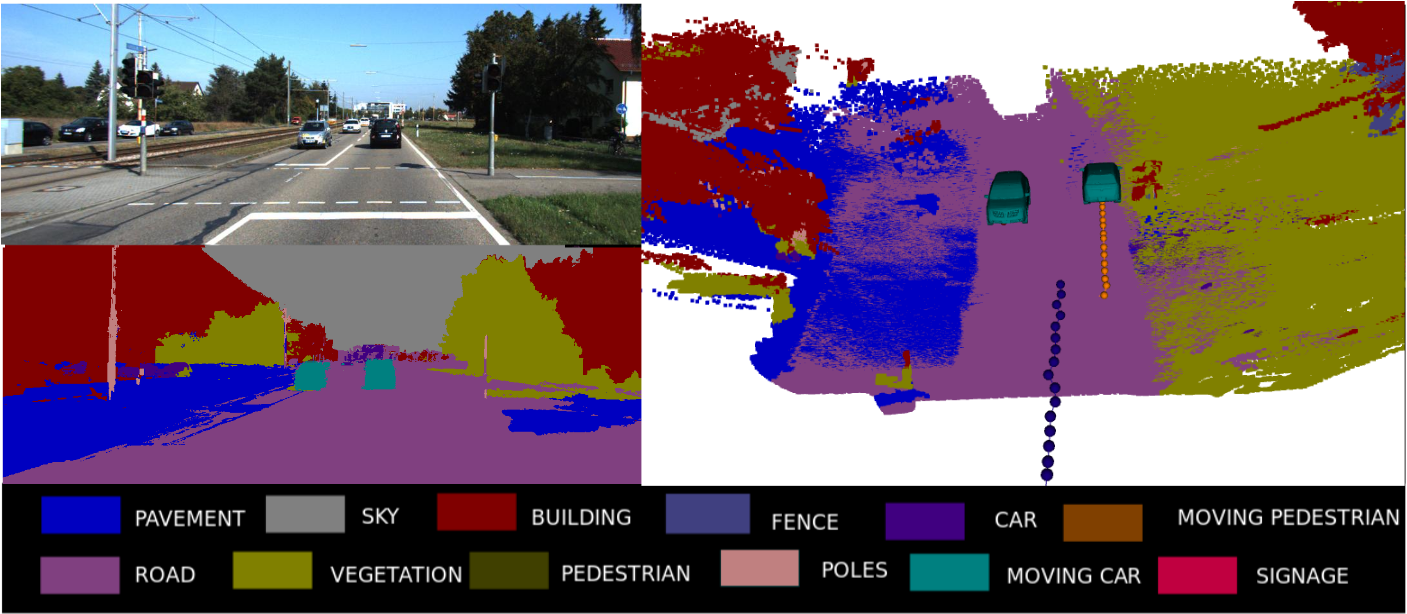}
\caption{
\small
{\bf Overview of our approach}:  {\bf Top left} A frame from highway sequence of the KITTI dataset has been used as input. {\bf Bottom left} Semantic Motion Segmentation to provide both motion and semantic understanding of the scene. {\bf Right} 3d reconstruction of the semantic map on the highway sequence with the trajectories of the moving objects overlaid. (Best viewed in color)
}
\label{fig:pipeline1}
\end{figure}

In urban environments, vSLAM is challenging particularly because of the presence of
\emph{dynamic objects}. Indeed, it is difficult to capture videos of a city without observing moving
objects like cars or people. However, dynamic objects are a source of error in vSLAM systems,
since the basic components of such algorithms make the fundamental assumption that the
world being observed is static. While optimization algorithms are designed to handle random noise
in observations, dynamic objects are a source of \emph{structured} noise since they do not
conform to models of random noise distributions (like Gaussian distributions, for example).
To overcome such difficulties, RANSAC based procedures for camera pose estimation and
3D reconstruction have been developed in the past, which treat dynamic objects as outliers and
remove them from the reconstruction process.

While successful attempts have been made to isolate and discard dynamic objects
from such reconstruction processes, there are several recent applications that \emph{benefit}
from \emph{reconstructions} of such objects. For example, reconstructing dynamic urban
traffic scenes are useful since traffic patterns can be studied to produce autonomous
vehicles that can better navigate such situations. Reconstructing dynamic objects are also
useful in indoor environments when robots need to identify and avoid moving obstacles in 
their path~\cite{sun3d}. 

\begin{figure*}[!th]
\centering
\includegraphics[width=170mm,height=52mm]{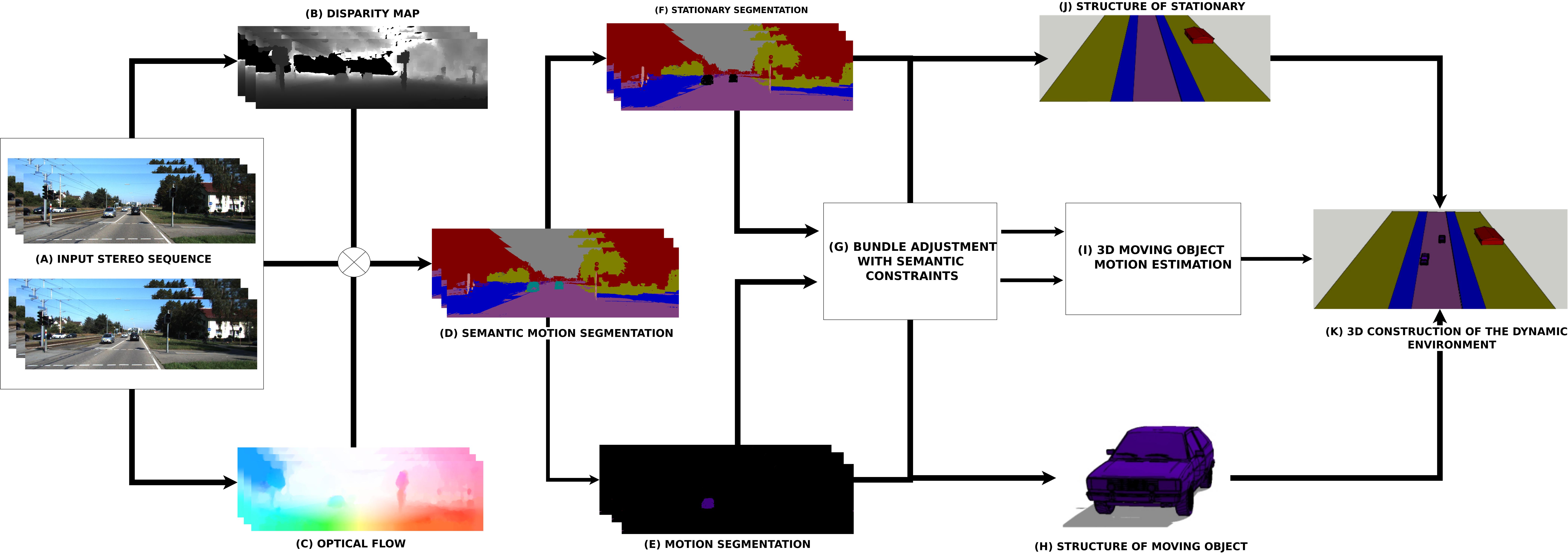}
\caption{
\small
{\bf Illustration of the proposed method}. The system takes a sequence of rectified stereo images (A). Our formulation computes the 
semantic motion segmentation (D) using the depth(B) and optical flow(C) information. We segment the moving objects (E) from the stationary background (F). 
  We compute accurate structure of the 
static background (J) and the moving object (H) with the help of bundle adjustment (G). This leads to state-of-the-art 3d reconstruction of the dynamic environment(K) with the help of moving object trajectory estimation(I). 
(Best viewed in color.)
}
\label{fig:pipeline}
\end{figure*}  
    
Reconstructing dynamic objects in videos present several challenges. Firstly, moving objects
in images and videos have to be segmented and isolated, before they can be reconstructed. This
in itself is a challenging problem in the presence of image noise and scene clutter. Degeneracies
in camera motion also prevent accurate motion segmentation of such objects.
Secondly, upon isolation, a separate vSLAM procedure must be initialized for \emph{each} moving object,
since objects like cars often move independent of each other and thus have to be treated as such.
Often moving objects like cars occupy only a small portion of the image space in a video
(Figure~\ref{fig:pipeline1}), because of which
dense reconstructions are infeasible since getting long accurate feature correspondence
tracks for such objects is difficult. Absence of large number of feature correspondences also
hinders accurate estimation of the car's pose with respect to a world coordinate system.
Finally, such objects cannot be reconstructed in isolation
from the static scene, since optimization algorithms like bundle adjustment do not preserve
contextual information like the fact that the car must move along a direction perpendicular to 
the normal of the road surface.

In this paper, we look at the problem of dynamic scene reconstruction. We present an end-to-end
system that takes a video, segments the scene into static and dynamic components and 
reconstructs \emph{both} static and dynamic objects separately. Additionally, while reconstructing
the dynamic object, we impose several novel constraints into the bundle
adjustment refinement that deal with noisy feature correspondences, erroneous 
object pose estimation, and contextual information. To be precise, we propose the following
contributions in this paper
\begin{itemize}
  \item We use a \emph{new semantic motion segmentation} algorithm using multi-layer dense 
  CRF which provides state-of-the-art motion segmentation and object class labelling.
\item We incorporate \emph{semantic contextual information} like support relations between the road
  surface and object motion, which helps better localize the moving object's pose vis-a-vis the
  world coordinate system, and also helps in reconstructing them.
\item We describe a \emph{random sampling strategy} that enables us to maintain the feasibility
  of the optimization problem in spite of the addition of a large number of variables.
\end{itemize} 
 
We evaluate our system on 4 challenging KITTI Urban tracking datasets captured using a stereo camera.
We are able to achieve average relative error reduction by 41.58 \% for one sequence based on 
Absolute Trajectory Error (ATE) for Root Mean Square Error relative to VISO2 ~\cite{lenz}, while
we get an improvement of 13.89 \% relative to traditional bundle adjustment.

This paper is organized as follows. We cover related work in Section~\ref{sec:RW}, and present
a system overview in Section~\ref{sec:3D}.
We describe process of motion segmentation using object class semantic constraints 
in Section \ref{sec:SMS}. We track and initialize multiple  moving bodies which we then 
optimize using a novel bundle adjustment in Section \ref{sec:TE}. Finally we show experimental
results on challenging datasets in Section~\ref{sec:exp}, and conclude in Section~\ref{sec:con}.

\section{\bf RELATED WORK}
\label{sec:RW}
Our system involves several components like semantic motion segmentation,
dynamic body reconstruction using multibody SLAM, and trajectory optimization.
We focus on each one of our components and draw references to relevant
works in the literature in this section.
\subsection{Dynamic body reconstruction}
Dynamic body reconstruction is a relatively new development in 3D reconstruction
with sparse literature on it. The few solutions 
in the literature can be categorized into decoupled 
and joint approaches.  Joint approaches like \cite{roussos2012dense}  
use monocular cameras to jointly estimate the depth maps, do motion segmentation and 
motion estimation of multiple bodies. Decoupled approaches like  
\cite{YuanM06} \cite{MulVSLAM_Abhijit_ICCV2011}  have a 
sequential pipeline where they segment motion and independently reconstruct the moving and 
static scenes. Our approach is a decoupled approach but essentially differs from other 
approaches, as we use a novel algorithm for semantic motion segmentation which is 
leveraged to obtain accurate localization of the moving objects through smoothness and 
planar constraints to give an accurate semantic map.  

\subsection{Semantic motion segmentation}
 Semantics have been used extensively for reconstruction \cite{Sengupta_icra_2013}  
\cite{conf/cvpr/ValentinSWST13} \cite{conf/cvpr/HaneZCAP13} but haven't been exploited 
in motion segmentation till recently~\cite{icvgippaper}.
Generally, motion segmentation has been approached using geometric constraints~\cite{MulVSLAM_Abhijit_ICCV2011}
or by using affine trajectory clustering into subspaces~\cite{subspace2009}.
In our approach we use motion \emph{along} with semantic cues to segment the scene into static and dynamic objects,
which allows us to work with fast moving cars, occlusions and disparity failure. 
We show a typical result of the motion segmentation algorithm 
in (Figure \ref{fig:pipeline1})(bottom left) where each variable is 
labelled for both multi-variate semantic class and binary motion class.  
\subsection{Multi-body vSLAM}
In dynamic scenes, decoupled approaches have motion segmentation followed by tracking 
each independently moving object to perform vSLAM. Traditional SLAM approaches with single 
motion model fail in such cases, as moving bodies cause reconstruction errors. 
Our approach employs Multi Body vSLAM framework
\cite{MulVSLAM_Abhijit_ICCV2011}  where we propose a novel trajectory optimization to 
with semantic constraints to show dense reconstruction results of moving objects. 

\subsection{Semantic constraints for reconstruction}
Recent approaches to 3D reconstruction have either used semantic information in a qualitative
manner~\cite{Sengupta_icra_2013}, or have only proposed to reconstruct indoor scenes using
such information~\cite{sun3d}. Only Yuan \etal~\cite{YuanM06} propose to add
semantic constraints for reconstruction. While our approach is similar to theirs, 
they use strict constraints for motion segmentation without regard to appearance information
whereas our approach works for more general scenarios as it employs a more powerful
inference engine in the CRF.

\section{\bf SYSTEM OVERVIEW}
\label{sec:3D}
We give an illustration of our system in Figure~\ref{fig:pipeline}. Given rectified input images
from a stereo camera, we first compute low level features like SIFT descriptors, optical flow
(using DeepFlow~\cite{weinzaepfel:hal-00873592}) and stereo~\cite{urtasun14eccv}. These are then used to compute semantic motion
segmentation, as explained in Section~\ref{sec:SMS}. Once semantic segmentation is done per image,
we isolate stationary objects from moving objects and reconstruct them
independently. To do this, we connect moving objects across frames into
tracks by computing SIFT matches on dense SIFT features~\cite{vedaldiVLFeat}. 
Then we perform camera resectioning using EPnP~\cite{lepetit} for stationary and ICP for moving objects,
to register their 3D points across frames.
This is then followed by bundle adjustment with semantic constraints (Section~\ref{sec:TE}), 
where we make use of the semantic and motion labels assigned to the segmented scene
to obtain accurate 3D reconstruction. We then fuse the stationary and moving object
reconstructions using an algorithm based on the truncated signed distance function
(TSDF)~\cite{ZhouICCV2013Elastic}. Finally, we transfer labels from 2D images to 3D data
by projecting 3D data onto the images, and using a winner-takes-it-all approach to assign
labels to 3D data from the labels of the projected points.

\section{\bf SEMANTIC MOTION SEGMENTATION}
\label{sec:SMS}
In this section, we deal with the first module of our system. 
A sample result of our segmentation algorithm is shown in Figure~\ref{fig:pipeline1}. With
input images from a stereo camera, we give an overview on how we perform 
semantic segmentation \cite{LadickyRKT09} to first separate
dynamic objects from the static scene. We combine classical semantic segmentation with
a new set of motion constraints proposed in \cite{icvgippaper} to perform semantic 
motion segmentation, that \emph{jointly} optimizes for semantic and motion segmentation.
While we give an overview of the formulation in this section, for brevity, methodologies
used for training, testing and the rationale behind using mean field approximations is outlined
in~\cite{icvgippaper}.


We do joint estimation of motion and object labels by exploiting the fact that they
are interrelated. We formulate the problem as a joint 
optimization problem of two parts, object class segmentation and motion segmentation. We 
define a dense CRF where the set of random variables $Z=\lbrace Z_1,Z_2,....,Z_N \rbrace$  
corresponds to the set of all image pixels $i \in \mathcal{V} = \lbrace 1,2,...,N\rbrace$. 
Let 
$\mathcal{N}_i $ denote the neighbors 
of the variable $Z_i$ in image space. Any possible assignment of labels to the random variables will be 
called a labelling and denoted by $z$. We define the energy of the joint CRF as
\begin{align}
E^\mathcal{J}  (z)=\sum_{i \in \mathcal{V}} \psi^\mathcal{J}_i (z_i) +\sum_{i \in \mathcal{V} , j \in \mathcal{N}_i} \psi^\mathcal{J}_{i,j} (z_i,z_j)
\end{align}
where $\psi^\mathcal{J}_i$ is the joint unary potential and $\psi^\mathcal{J}_{i,j}$ represents
the joint pairwise potential. We describe these terms in brief in the next two sections.
\subsection{\bf Joint  Unary Potential:}
The  joint unary potential $\psi^\mathcal{J}_i$ is defined as an interactive potential 
term which incorporates a relationship between the object class and the corresponding motion 
likelihood for each pixel. Each random variable $Z_i$ = [$X_i$,$Y_i$] 
takes a label $z_i$ = [$x_i$,$y_i$], from the product space of object class and motion labels.
The combined unary potential of the joint CRF is
\begin{align}
\psi_{i,l,m}^\mathcal{J}([x_i,y_i])= \psi_i^O(x_i)+ \psi_i^\mathcal{M}(y_i) + \psi_{i,l,m}^{O \mathcal{M}}(x_i,y_i)
\end{align}
The object class unary potential $\psi^O_i (x_i)$ describes the cost of the pixel taking the 
corresponding label and is computed using pre-trained models of color, texture and 
location features for each object as in \cite{Shotton06textonboost}. The new motion class unary 
potential $\psi^\mathcal{M}_i (y_i)$ is given by the motion likelihood of the pixel and is computed 
as the difference between the predicted and the measured optical flow. The measured flow is computed 
using dense optical flow. The predicted flow measures how much the object needs to move given its
depth in the current image and assuming it is a stationary object. Objects deviating from the 
predicted flow are likely to be dynamic objects. It is computed as
\begin{align}
   \hat{X}'=KRK'X+KT/z
     \label{eq:motion_estimate}
\end{align}
where K is the intrinsic camera matrix, R and T are the translation and rotation of 
the camera respectively and z is the depth~\cite{icvgippaper}. X is the location of the pixel in image coordinates 
and $\hat{X}'$ is the predicted flow vector of the pixel given from the motion of the camera. 
Thus the unary potential is now computed as
\begin{align}
  \psi^\mathcal{M}_i (x_i) =({ (\hat{X}'-X')^T \Sigma^{-1}(\hat{X}'-X') })
  \label{eq:motion}
\end{align}
where $\Sigma$ is the sum of the covariances of the predicted and measured flows 
as shown in \cite{Romero-CanoN13}, \& $\hat{X}'-X'$ represents 
the difference of the predicted flow and measured flow. The object-motion unary potential 
$\psi^{O \mathcal{M}}_{i,l,m} (x_i,y_j)$ incorporates the object-motion class compatibility and 
can be expressed as
\begin{align}
  \psi^{O \mathcal{M}}_{i,l,m} (x_i,y_j) =    \lambda(l,m)
    \label{eq:unary_joint}
\end{align}
where $\lambda(l,m) \in [-1,1]$ is a learnt correlation term between the motion and object 
class label. $\psi^{O \mathcal{M}}_{i,l,m} (x_i,y_j)$ helps in incorporating the relationship between 
an object class and its motion (for example, trees and roads are stationary, but cars move).
We use a piecewise method for training the label and motion correlation matrices 
using the modified Adaboost framework~\cite{icvgippaper}, as described in~\cite{icvgippaper}.

\subsection{\bf Joint  Pairwise Potential:}
The joint pairwise potential $\psi_{ij}^\mathcal{J}(z_i,z_j)$ enforces the consistency of 
object and motion class between the neighboring pixels. We compute the joint pairwise potential as
\begin{align}
\psi^\mathcal{J}_{ij}([x_i,y_i],[x_j,y_j]) =
 \psi_{ij}^O(x_i,x_j)
 + \psi_{ij}^\mathcal{M}(y_i,y_j)
\end{align}
where we disregard the joint pairwise term over the product space. The object class pairwise potential 
takes the form of a Potts model
\begin{align}
  \psi^O_{i,j} (x_i,x_j) =\left\{ \begin{array}{ll}
         0 & \mbox{if ${x_i=x_j}$}\\
        p(i,j) &  \mbox{if $x_i \neq x_j$}\end{array} \right. \
\end{align}
where $p(i,j)$ is given as the standard pairwise potential as given in
\cite{Krahenbuhl_Koltun_2011}.

The motion class pairwise potential $\psi^\mathcal{M}_{i,j} (y_i,y_j) $ is given as the 
relationship between neighboring pixels and encourages the adjacent pixels in the image to 
have similar motion label. The cost of the function is defined as
\begin{align}
  \psi^\mathcal{M}_{ij} (y_i,y_j) =\left\{ \begin{array}{ll}
         0 & \mbox{if ${y_i=y_j}$}\\
        g(i,j) &  \mbox{if $y_i \neq y_j$}\end{array} \right. \
\end{align}
where $g(i,j)$ is an edge feature based on the difference between the flow of the neighboring pixels
($g(i,j) = |f(y_i)-f(y_j)|$)
\& $f(\cdot)$ is returns the flow of the corresponding pixel.

\subsection{\bf Inference and learning:} 
We follow Krahenbuhl et al {\cite{Krahenbuhl_Koltun_2011}} to perform inference on this
dense CRF using a mean field approximation. In this approach we try to find a mean field approximation $Q(z)$ 
that minimizes the KL-divergence $D(Q \Vert P)$ among all the distributions $Q$ that can be 
expressed as a product of independent marginals, $Q(z)=\prod_i Q_i (z_i)$. We can further 
factorize $Q$ into a product of marginals over multi-class object and binary motion 
segmentation layer by taking $Q_i (z_i)=Q^O_i(x_i) Q_i^\mathcal{M}(y_i)$. 
Here $Q^O_i$ is a multi-class distribution over the object labels, and $Q_i^\mathcal{M}$ is a 
binary distribution over moving or stationary classes ($ Q_i^\mathcal{M} \in \lbrace 0,1 \rbrace$). We compute inference 
separately for both the layers i.e object class layer and motion layer~\cite{icvgippaper}.

\section{\bf TRAJECTORY ESTIMATION} 
\label{sec:TE}

The motion segmented images of the static world and moving objects are then used as input to localize 
and map each object independently. In this section, we propose a novel framework for trajectory computation 
for static or moving objects from a moving platform. The below process is carried out for all the moving 
objects and the camera mounted vehicle\footnote{ Henceforth referred as camera}. Let us introduce 
some preliminary notations for trajectory computation. The extrinsic parameters for frame  $k=1,2,3,4...n$ 
are the rotation matrix $R_k$ and the camera center $C_k$ relative to a world coordinate system. 
Then the translation vector between the world and the camera coordinate systems is $T_k$ = -$R_k C_k$ .

  
\paragraph{\bf Trajectory  Initialization}
 We initialize the motion of each object separately using SIFT feature points. SIFT feature points are 
 tracked using dense optical flow between consecutive pair of frames. Key points with valid depth 
 values are used in a 3-point-algorithm within a RANSAC framework to find the robust relative 
 transformation between pairs of frames. We obtain pose estimates of the moving object in the world 
 frame by chaining the relative transformations together in succession. 
 
 For moving objects the initial frame k where detection occurs is taken as the starting point. 
 Trajectory estimates are then initialized for each object independently corresponding to the frame k 
 assuming the camera is static.  

\paragraph{\bf 3D Object Motion Estimation} 

Once 3D trajectories are estimated for each object independently, we need to map these trajectories
onto the world coordinate system. Since, we are dealing with stereo data and for every frame we have
3D information, this mapping can be represented as simple coordinate transformations. Also, since we
are not dealing with monocular images, the problem of relative scaling can be avoided. 

Given the pose of the real camera in the $k^{th}$ frame ($(R_k^c,T_k^c)$) 
and virtual camera $(R_k^v,T_k^v)$~\cite{YuanM06} computed during trajectory initialization 
described earlier, we should be able to compute the pose of the $b^{th}$ object $(R_k^b,T_k^b)$ 
relative to its original position in the first frame
in the world coordinate system. The object rotation $R_k^b$ and translation $T_k^b$ are given as
 \begin{equation}
R_k^b=  (R_k^c)^{-1} R_k^v, \qquad T_k^b=  (R_k^c)^{-1} (T_k^v-T_k^c)
\end{equation}
Thus we get the localization and sparse map of both the static and moving world. We found this approach 
to object motion estimation to be better on both small and long sequences than VISO 2~\cite{lenz}.
 
\begin{figure}
  \centering
  \includegraphics[width=65mm,height=45mm]{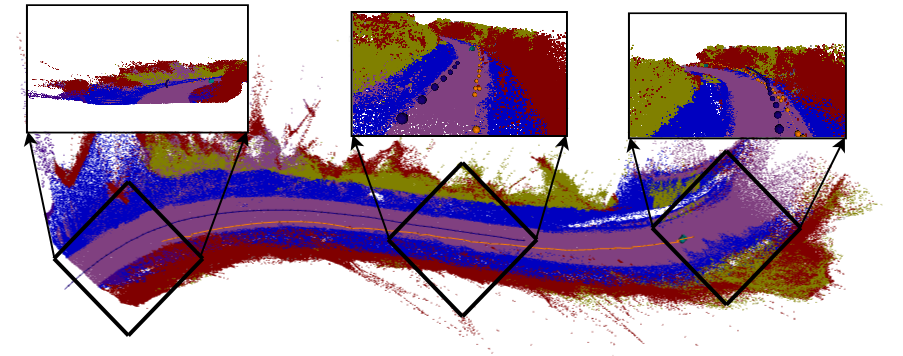}
  \caption{ \small Reconstruction result for \textbf{KITTI 4} sequence. Note the accurate reconstruction of trajectories and
  of the car and the camera. Please see supplementary video for
  further details.}
\end{figure} 
 
\subsection{\bf Bundle Adjustment}  \label{sec:T} 
Once 3D object motion and structure initialization has been done, we need to refine the structure and
motion using bundle adjustment (BA). In this section, we describe our framework for BA to refine the trajectory and sparse
3D point reconstruction of dynamic objects along with \emph{several
novel} constraints added to BA that increase the accuracy of our trajectories and 3D points. 
We term these constraints \emph{semantic or contextual} constraints since they represent our
\emph{understanding} of the world in a geometric language, which we use to effectively optimize
3D points and trajectories in the presence of noise and outliers.  The
assumptions underlying these constraints derive from \emph{commonly} observed \emph{shape} and \emph{motion} traits of 
cars in urban scenarios. For example the normal constraints follow the logic that the motion of a 
dynamic object like a vehicle is always on a plane (the road surface) and hence constrained by its normal.
Similarly, the 3D points on a dynamic object are constrained to lie within a 3D ``box'' since dynamic objects
like cars cannot be infinitely large. Finally, our trajectory constraints encode the fact that dynamic
objects have smooth trajectories, which is often true in urban scenarios.
In summary, we try to minimize the following objective function
\begin{eqnarray}
min\sum_\mathrm{i}\sum_{p\in V(i)} \mathtt{BA2D} + \lambda \mathtt{BA3D} + \lambda  \mathtt{TC} +
\mathtt{NC} + \mathtt{BC}
\label{eqn:baopt}
\end{eqnarray} 
where $\mathtt{BA2D}$ represents the 2D BA reprojection error
($\Vert \tilde{x}_p^i - K[R_i~\vert~T_i]X_p\Vert ^2$), $\mathtt{BA3D}$ represents the 3D registration error
common in optimization over stereo images ($\Vert \tilde{X}_p^i-[R_i~\vert~T_i]X_p\Vert ^2$) and
$\mathtt{TC,~NC,~BC}$ represent various optimization terms that can be seen as 
imposed constraints on the resulting shape and trajectories as explained below. Here $i$ indexes into images,
and $\tilde{}$ represents variables in the camera coordinate system, with other quantities being
expressed in the world coordinate system. Also, $p \in V(i)$ represents pixels visible in image $i$.


\subsubsection{\bf Planar Constraint} We constrain motion to be perpendicular to the ground plane 
where the ground plane normal is found from the initial 3D reconstruction of the ground.  
\begin{eqnarray}
  \mathtt{NC1}: & & N_{g}\cdot T_{c}
\end{eqnarray}
where $N_{g}$ is the normal of the ground plane, $t_{c}$ is the direction of translation
in the world coordinate system. Since 3D reconstruction of the ground can be noisy, estimation of
$N_{g}$ is done using least squares. Alternatively, we could follow a RANSAC based framework of
selecting $m$ top hypotheses for the normal $N_g^i$ ($i = 1\dots m$), and allow bundle adjustment
to minimize an average error of the form
\begin{eqnarray}
  \mathtt{NC2}: & & \sum_{i=1}^m N_g^i \cdot T_{c}
\end{eqnarray}

\subsubsection{\bf Smooth Trajectory Constraints} We enforce smoothness in trajectory, a valid
assumption for urban scenes, by constraining camera translations in consecutive frames as
\begin{eqnarray}
  \mathtt{TC1}: & & \Vert (T_c^{k} - T_c^{k-1}) \times T_c^{k})\Vert
\end{eqnarray}
where $T_c^{k},T_c^{k-1}$ are the 3d translations at frame k and k-1. Alternatively, we could also
minimize the norm between two consecutive translations unlike $\mathtt{TC1}$, which only penalizes
direction deviations in translation.
\begin{eqnarray}
  \mathtt{TC2}: & & \Vert (T_c^{k+1} - 2*T_c^{k} + T_c^{k-1}) \Vert^2
\end{eqnarray}

\subsubsection{\bf Box Constraints} Depth estimation of objects like cars are generally noisy
because their surface is not typically Lambertian in nature, and hence violates the basic
assumptions of brightness constancy across time and viewing angle. Furthermore, noise in depth
infuses errors into the estimated trajectory through the trajectory initialization component.
To improve the reconstruction accuracy in such cases, and to limit the destructive effect that
noisy depth has on object trajectories, we introduce shape priors into the BA cost function that
essentially constrains all the 3D points belonging to a moving object to remain with a ``box''.KITTI
More specifically, let $X^b_i \& X^b_j$ be two 3D points on a moving object $O^b$. For every 
such pair of points on the object, we define the following constraint

\begin{eqnarray}
  \mathtt{BC1}: & & \sum_{\forall {i, j}} \Vert X^b_i - X^b_j - B(i, j) \Vert^2 \label{eqn:origbox} \\
              & & -\mathbf{\delta} \le B(i, j) \le \mathbf{\delta}                  \nonumber 
\end{eqnarray}
where $B(i, j)$ is a vector of bounds with individual components ($b_x(i, j), b_y(i, j), b_z(i, j)$)
and $\delta$ is a vector of positive values. 

Note that the above equation is defined for every pair of points on the object, which leads to
a {\it quadratic explosion} of terms since $B(i, j)$ is a separate variable for each pair. 

\paragraph{\bf Alternate Formulations}
One way to reduce the explosion would be to reduce the number of variables added because of the box
constraints to BA. This could be done by alternatively minimizing the following terms instead
of the constraint in equation~(\ref{eqn:origbox})
\begin{eqnarray}
  \mathtt{BC2}: & & \sum_{\forall (i, j)} \Vert X^b_i - X^b_j - b(i, j) \Vert^2, -\delta \le b(i, j) \le \delta \label{eq:bc2} \\ 
  \mathtt{BC3}: & & \sum_{\forall (i, j)} \Vert X^b_i - X^b_j - B \Vert^2, -\mathbf{\delta} \le B \le \mathbf{\delta} \label{eq:bc3} \\
  \mathtt{BC4}: & & \sum_{\forall (i, j)} \Vert X^b_i - X^b_j - b \Vert^2, -\delta \le b \le \delta  \label{eq:bc4}
\end{eqnarray}
where $b(i, j)$ in equation~(\ref{eq:bc2}) is a scalar common to all 3 dimensions, $B$ (equation~(\ref{eq:bc3})) is a $3 \times 1$ vector common to all
point pairs, and $b$ (equation~(\ref{eq:bc4})) is a scalar common to all pairs and dimensions.

\paragraph{\bf Alternate Minimization Strategies}
It is now known that a lot of information in terms like $\mathtt{BC1, BC2, BC3, BC4}$ above
are redundant in nature~\cite{siddharth15icra}, and there is essentially a small ''subset'' of pairs
which is sufficient to produce optimal or near-optimal results in such cases. However, 
it is not clear how to pick this small subset. Here, we take the help of the Johnson-Lindenstrauss
theorem and its variants~\cite{matousek,dasgupta,clarkson}, to select a random set of pairs from the ones available, such that we closely approximate
the $\mathtt{BC}$ error when all the point pairs are used.

More specifically, the terms expressed in $\mathtt{BC1, BC2, BC3, BC4}$ can all be
expressed in the form
\begin{eqnarray}
  \mathtt{BCLin}: & & \Vert A X - B \Vert^2
\end{eqnarray}
where $X$ is a concatenation of all 3D points, and $B$ is a collection of all
car bounds. The matrix $A$ is constructed in such a way that each row of
$A$ consists of only two non-zero elements at the $i^{th}$ and $j^{th}$ positions
with values $1$ and $-1$ respectively, and they represent the difference $X^b_i - X^b_j$.
Note that the dimensions of $A$ are of the order $3\comb{n}{2} \times 3n$, where
$n$ is the number of 3D points.
Notice that for $n=3000$, $\comb{n}{2}$ is approximately $4.5$ million, and is highly slow to optimize!
To reduce this computational burden, we embed the above optimization problem in a randomly selected
subspace of considerably lower dimension, with the guarantee that the solution obtained in the subspace
is close to the original problem solution with high probability.
To do this, we draw upon a slightly modified version of the \emph{affine embedding} theorem presented in~\cite{clarkson} which
states
\begin{theorem}
  For any minimization of the form $\Vert A X - B \Vert$, where $A$ is of size $m \times n$
  and $m \gg n$, there exists a \emph{subspace embedding matrix} $S : \mathbb{R}^m \mapsto \mathbb{R}^t$
  where $t = poly( n/\epsilon )$ such that
  \begin{equation}
    \Vert S A X - S B \Vert_2 = ( 1 \pm \epsilon ) \Vert A X - B \Vert_2 
  \end{equation}
  Moreover, the matrix $S$ of size $t \times m$ is designed such that each column of $S$ has
  only 1 non-zero element at a randomly chosen location, with value $1$ or $-1$ with equal 
  probability.
\end{theorem}

Note that since elements of $S$ are randomly assigned 1 or -1, the above transformation cannot
be exactly interpreted as a random sampling of pairs of points. However for the sake of
implementation simplicity, we ``relax`` $S$ to a random selection matrix. As we show later, 
empirically we get very satisfying results.

There can be several strategies to select random pairs of points for box constraints. We
experimented with the following in this paper.
\begin{itemize}
  \item $\mathtt{Strat1}$: Randomly select pairs from the available set.
  \item $\mathtt{Strat2}$: Randomly select one point, and create its pair with the 3D
    point that is farthest from the selected point in terms of Euclidean distance.
  \item $\mathtt{Strat3}$: Randomly select one point, and sort other points in descending order
    based on Euclidean distance with selected point. Pick the first point from the list that
    has not been part of any pair before.
\end{itemize}

%
%

Once the proper set of constraints are selected from the above choices, 
the final objective function in equation~\ref{eqn:baopt} is minimized with $L_{2}$ norm using CERES solver. \cite{ceres}. 



\begin{figure}
\begin{center}
  \centering
\includegraphics[width=85mm,height=35mm]{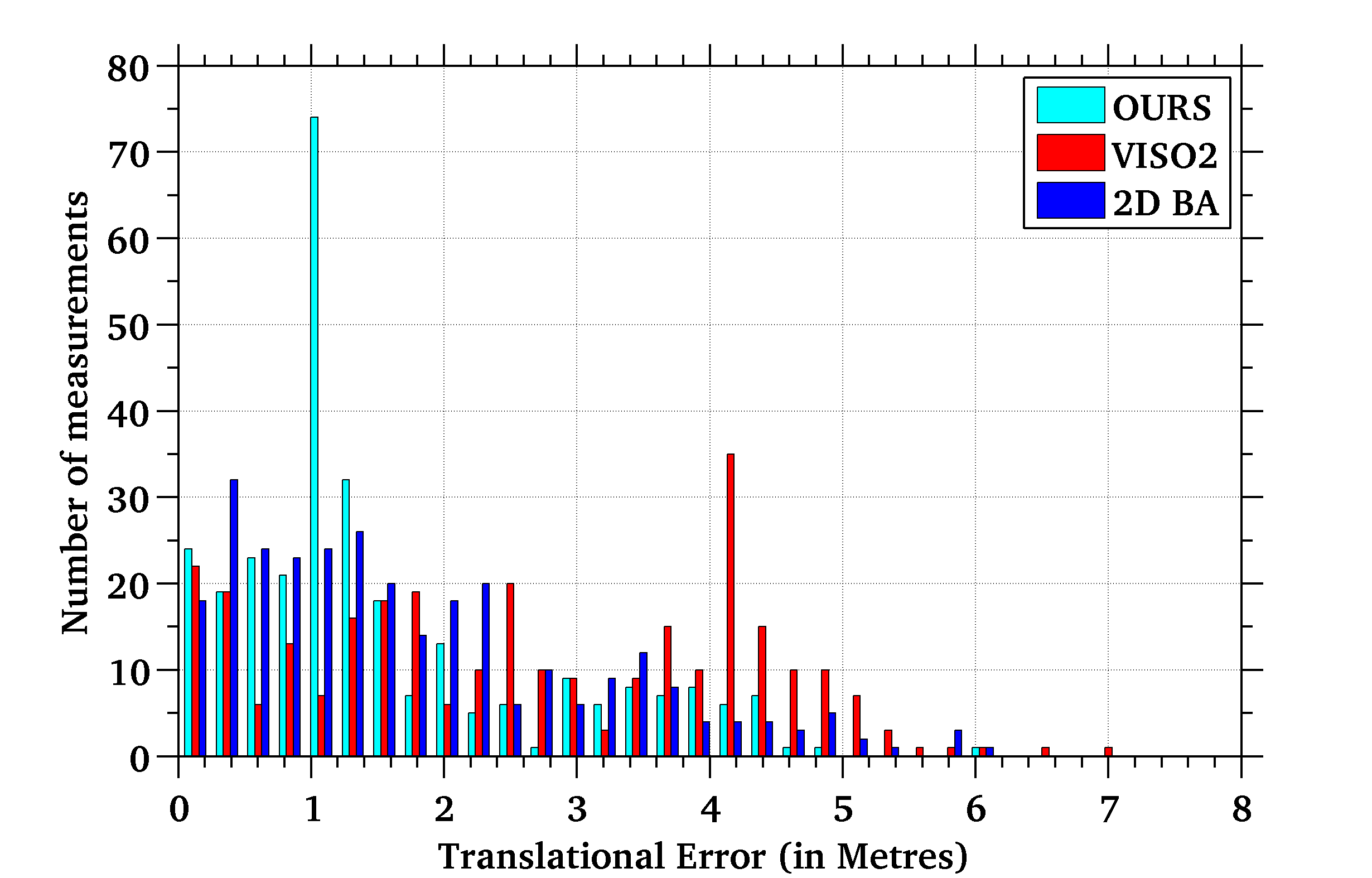} 
\end{center}
 \caption{
 \small
 Comparison of trajectory errors of our algorithm to VISO 2~\cite{lenz} and standard
 BA. The histogram plots RMSE magnitude on the x axis, and number of pose measurements in the trajectory 
 that fall within a particular range on the y axis. Note that most of our errors are 
 concentrated on the left (low error), while VISO 2~\cite{lenz} and BA are more evenly spread. 
The total summed error is: 2D-BA - 1.7919, VISO2 - 2.6185, OUR Approach - 1.5429.
}
\label{fig:histogram}
\end{figure} 

\section{\bf EXPERIMENTAL RESULTS}
\label{sec:exp}
In this section we provide extensive evaluation of our algorithms on both synthetic and real data.
For real datasets, we have used the KITTI tracking dataset for evaluation of the algorithm as the ground truth for 
localization of moving objects per camera frame is available. It consists of several sequences 
collected by a perspective car-mounted camera driving in urban,residential and highway 
environments, making it a varied and challenging real world dataset. We have taken four 
sequences consisting of 30, 212, 30, 100 images for evaluating our algorithm. 
We choose these 4 sequences as they pose serious challenges to the motion segmentation algorithm as the moving cars lie in the same subspace as the camera. 
These sequences also have a mix of multiple cars visible for short duration along with cars visible
for the entire sequence which allows us to test the robustness of our localization and reconstruction 
algorithms on both short and long sequences.  

\begin{table*}
\centering
\setlength\tabcolsep{1.5pt}

\begin{minipage}[b]{0.48\linewidth}
\begin{tabular}{|c|c|c|c|c|c|c|}
\hline
& Error Type & Without MS & MS & MS+Normal & MS+Normal \tabularnewline
&           &           &       &           &   +Trajectory
\tabularnewline
&           &           &       &           &
\tabularnewline
\hline
2D & rsme  &   1.416246 & 1.001566 & 0.941971 & 0.958505 \tabularnewline
\hline
2D & mean  & 1.212164 &    0.826189 & 0.764188 &  0.779054  \tabularnewline
\hline
2D & median  &  1.088891 &  0.677419 & 0.690825 & 0.716546 \tabularnewline

\hline
3D & rsme & 1.476649 &  0.959499 & 0.975747 &  0.978197  \tabularnewline
\hline
3D & mean  &   1.272985 & 0.786729 & 0.822169 & 0.824090 \tabularnewline
\hline
3D & median & 1.279508 & 0.712513 & 0.773672 & 0.769680  \tabularnewline

\hline
2D+3D & rsme &  1.472399 &  0.958505 &  0.958541 & 0.958541  \tabularnewline
\hline
2D+3D & mean & 1.269541 & 0.779054 & 0.779132 & 0.779132   \tabularnewline
\hline
2D+3D & median & 1.269238 &  0.716546 & 0.716967 & 0.716967  \tabularnewline
\hline
\end{tabular}
\caption{\small Static scene of the {\bf KITTI} dataset. The dataset has 212
frames. Note that adding Motion Segmentation (MS) 
drastically improves results.}
\label{tab:stationary_ate}
\end{minipage} \hfill
\begin{minipage}[b]{0.48\linewidth}
\setlength\tabcolsep{1.5pt}
\begin{tabular}{|c|c|c|c|c|c|c|}
\hline
& Error Type & MS & MS+Normal & MS+Normal &  MS+Normal+ \tabularnewline
&            &      &           & +Trajectory &       Trajectory+Box \tabularnewline
&           &       &           &           &(1000 constr) 
\tabularnewline
\hline
2D & rsme  & 2.425649 & 2.362224 & 2.351205 & \textbf{2.302849} \tabularnewline
\hline
2D & mean   &    1.989408 & 1.955466 &1.969793 & \textbf{1.937154} \tabularnewline
\hline
2D & median   &  1.669304 & 1.616398 & 1.685272 & \textbf{1.640389} \tabularnewline

\hline
3D & rsme  & 3.627977 & 3.587194 &  3.352087 & \textbf{3.270264} \tabularnewline
\hline
3D & mean   &  2.544718 & 2.527314 & 2.398578 &  \textbf{2.367702} \tabularnewline
\hline
3D & median  & 2.000463 & 1.997689 & 1.941246 &  \textbf{1.928450} \tabularnewline

\hline
2D+3D & rsme  &  2.357187 &  2.305733 & 2.296139 &  \textbf{2.254192} \tabularnewline
\hline
2D+3D & mean  & 2.035764 & 1.986784 & 1.971698  &  \textbf{1.881728} \tabularnewline
\hline
2D+3D & median  &  1.877257 & 1.759010 & 1.760857 &  \textbf{1.756554} \tabularnewline

\hline
\end{tabular}
\caption{\small Dynamic scene of {\bf KITTI} dataset of 212 frames. Note that adding box constraints over normal and trajectory
lead to the best results.}
\label{tab:KITTI2_ate}
\end{minipage}
%
\end{table*}
We do extensive quantitative evaluation on synthetic dataset. We generated 1000 3D points on a cube attached
to a planar ground to simulate a car and road. We then move the car over the road, while
simultaneously moving the camera to generate moving images after projection of the 3D points.
Finally we add Gaussian noise to both the 3D points on the car and the points on the road to
simulate errors in measurement. Correspondences between frames are automatically known as a result
of our dataset design.

\subsection{\bf Quantitative Evaluation of BA}
In this section, we do an extensive evaluation of the different terms proposed in
Section~\ref{sec:TE}. Note that we tried all the different terms and strategies proposed
here on real data as well, and in all cases conclusions derived from synthetic data
experiments are consistent with real data.

\subsubsection{\bf Evaluating Terms and Strategies}
\begin{figure}
  \centering
  \subfloat[]{\includegraphics[width=40mm,height=35mm]
  {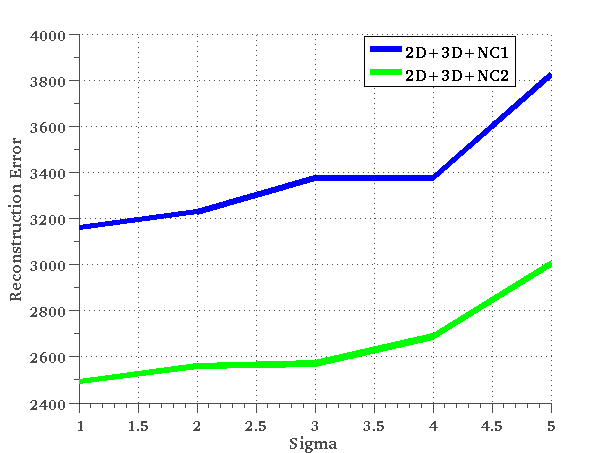}\label{fig:normal_constraint}}
  \subfloat[]{\includegraphics[width=40mm,height=35mm]
  {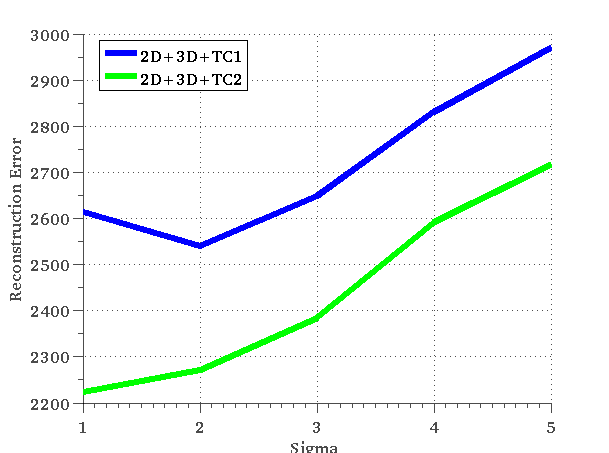}\label{fig:traj_constraint}}
  \caption{\small  Synthetic results for Normal and trajectory constraints. }
\end{figure}

In the following section we present the results for evaluation of various terms and strategies.
\paragraph{\bf Normal Constraint}
This constraint is a contextual constraint in the sense that it enforces the fact that 
the moving object is usually attached to a planar ground in urban settings, and so any deviation of
the object trajectory along the direction of the normal of the ground plane should be
penalized. While $\mathtt{NC1}$ computes a least-squares estimate for the normal which is optimal
under Gaussian noise, $\mathtt{NC2}$ computes several normal hypotheses using a RANSAC framework.
Figure (\ref{fig:normal_constraint}) shows the results comparing the two terms. We find that
$\mathtt{NC1}$ normally performs better.

\paragraph{\bf Trajectory Constraint}
The trajectory constraint enforces smoothness in moving object trajectories, by either enforcing
that the direction of motion should not change significantly between consecutive frames
($\mathtt{TC1}$) or enforcing that both direction and magnitude must be constrained
($\mathtt{TC2}$). Figure (\ref{fig:traj_constraint}) plots comparative results, and we infer that
$\mathtt{TC2}$ performs better.

\paragraph{\bf Box Constraint}
Box constraints enforce that the 3D reconstruction of the moving object in consideration must be
\emph{compact}. This is a useful constraint since gross errors in the depth of the object as
estimated by the stereo algorithm~\cite{urtasun14eccv} normally are not corrected by BA since
it settles into a local minima. Thus, to ``focus'' the BA towards better optimizing
the 3D structure, we add these constraints.
 
\begin{figure}
  \centering
  \subfloat[$\mathtt{BC}$ terms]{\includegraphics[width=40mm,height=35mm]{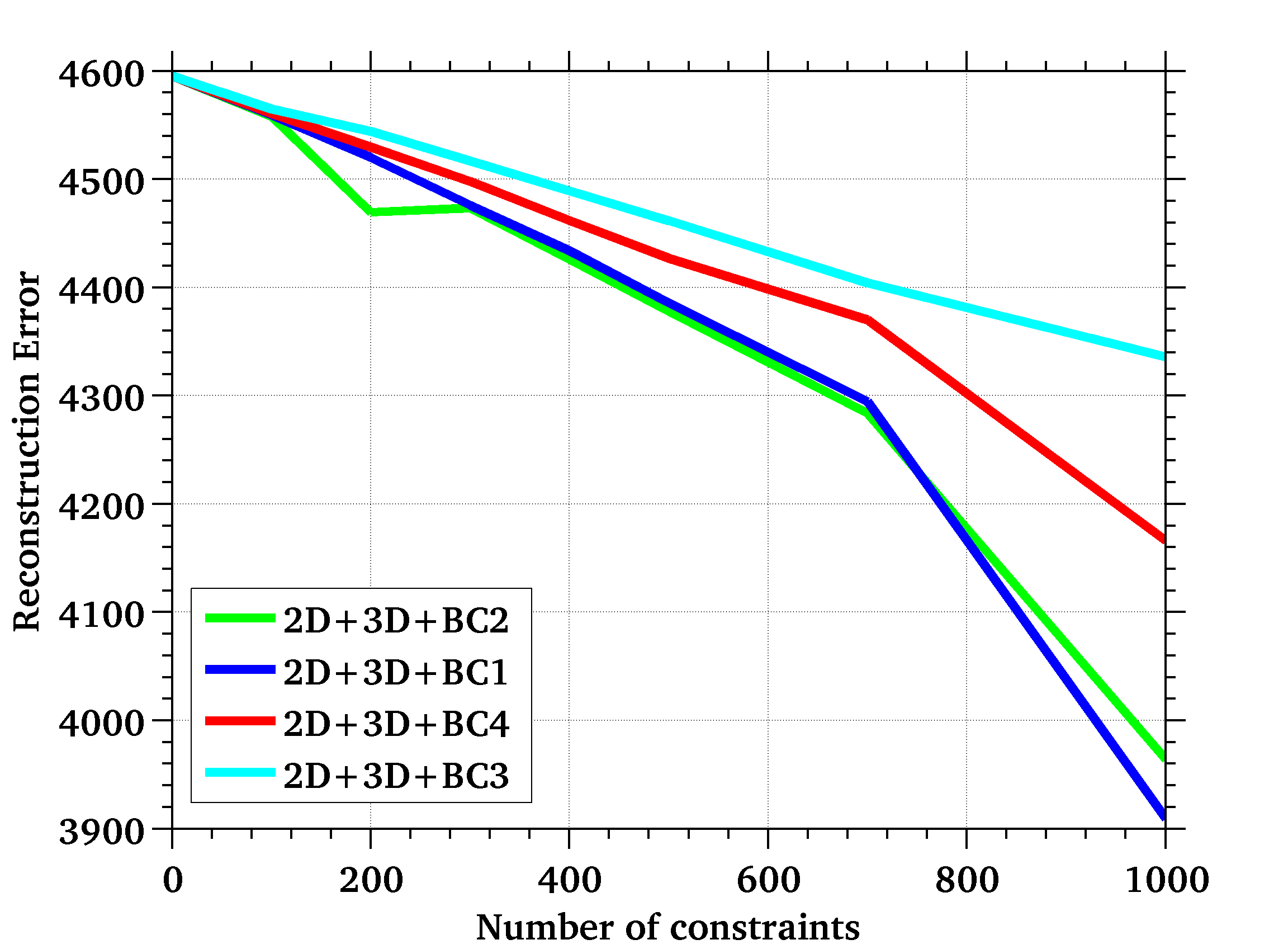}}
  \subfloat[$\mathtt{Strat}$ strategies]{\includegraphics[width=40mm,height=35mm]{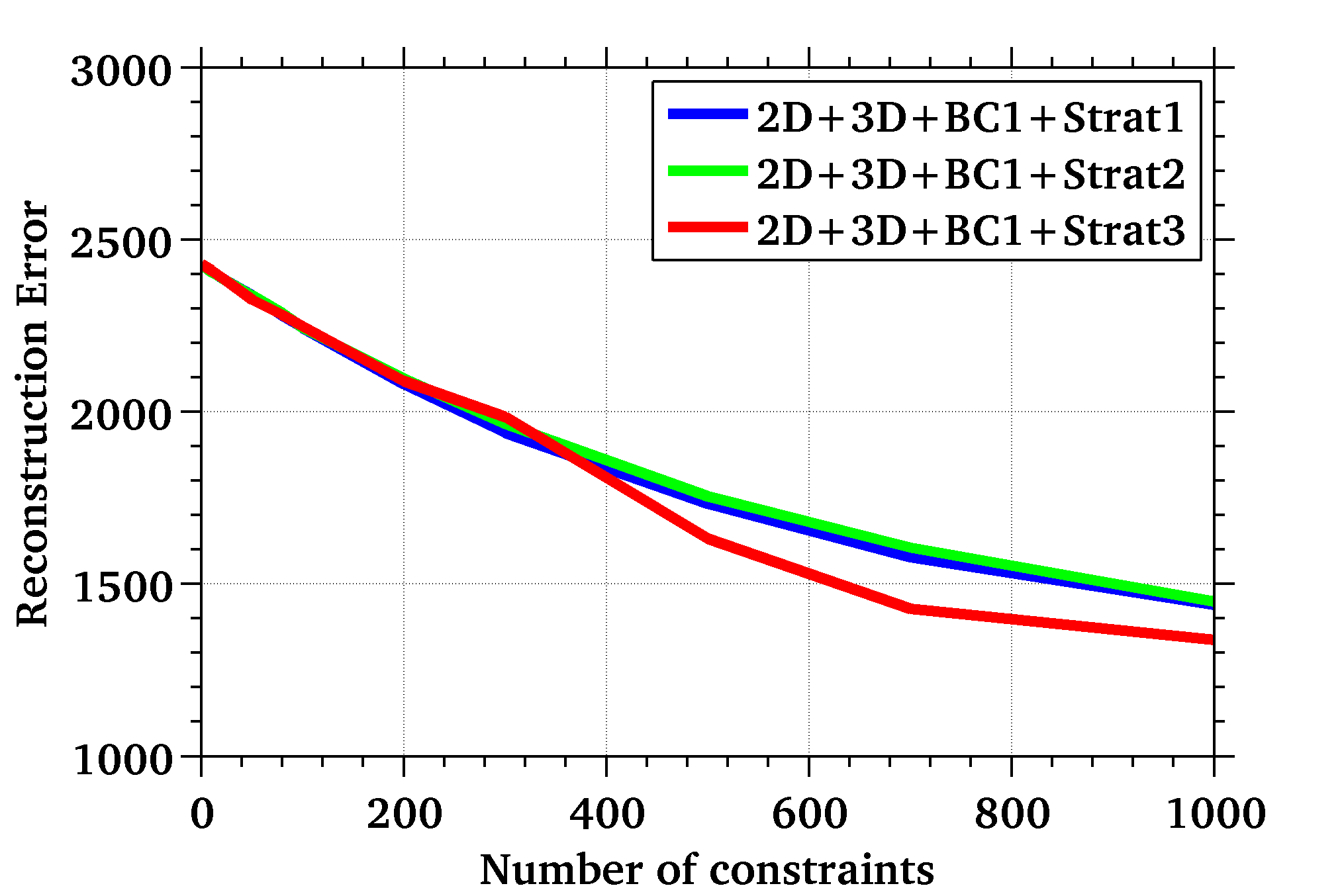}}
  \caption{\small Synthetic results for box constraints. Note that in the two experiments we added a large amount of noise
  and picked 1000 constraints from around 500000 pairs of points. While there isn't much difference between the
  terms in (a) as such, $\mathtt{Strat3}$ performs better than others in (b).}
  \label{fig:box_constraints}
\end{figure}

\paragraph{\bf Box Sampling Strategies}
Since box constraints lead to an explosion of terms added to BA, we experiment with 4 strategies
to reduce this computational burden by random sampling~\cite{matousek}.
Figure~(\ref{fig:box_constraints}) show results for various terms of box constraints, and various
strategies to optimize.  Normally we find that $\mathtt{BC1}$ along with $\mathtt{Strat3}$ performs
best.

\subsection{\bf Trajectory Evaluation}

\begin{figure*}[t!h]
\begin{center}

\begin{tabular}{c c c c}
 & \bf KITTI 1 & \bf KITTI 2 & \bf KITTI 3\\
\begin{sideways}\bf \centering  \ \    INPUT  \end{sideways} & 
  \includegraphics[width=52mm,height=15mm]{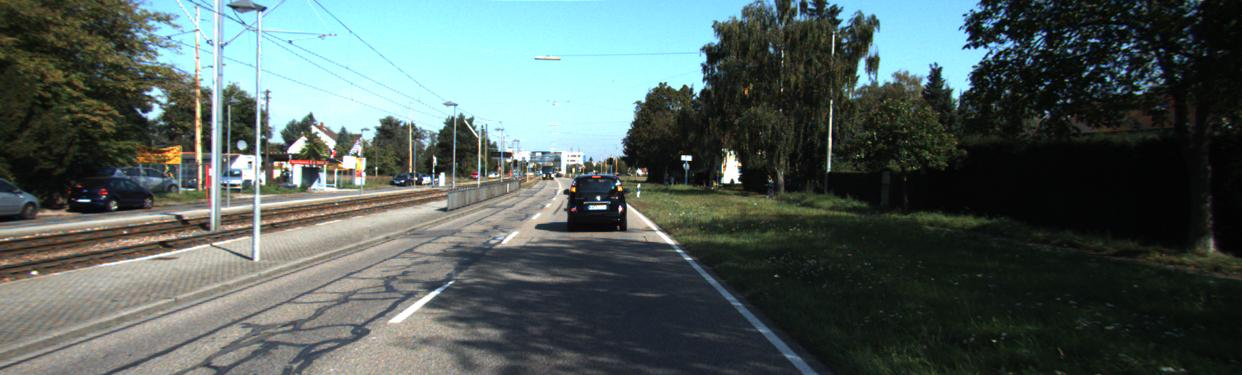} &
\includegraphics[width=52mm,height=15mm]{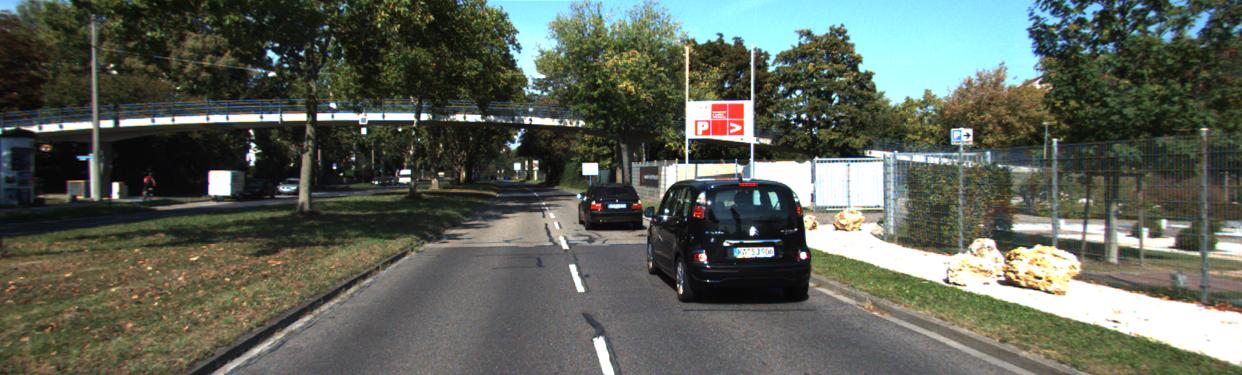} &
\includegraphics[width=52mm,height=15mm]{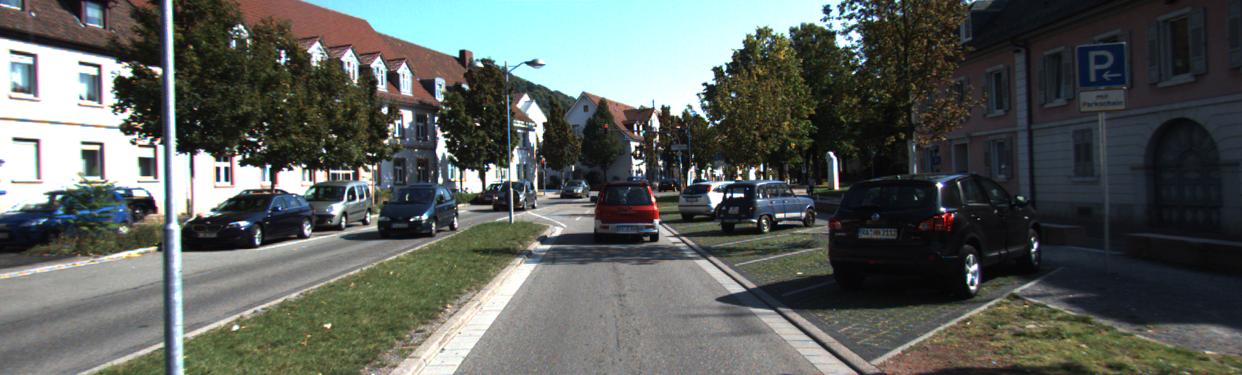}\\
 
\begin{sideways}\bf \centering  \ \     SMS \end{sideways} & 
\includegraphics[width=52mm,height=15mm]{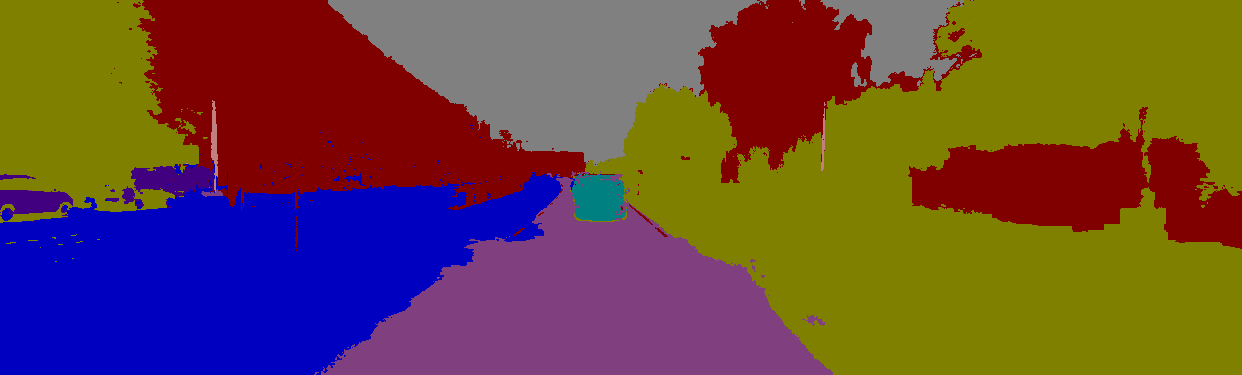} & 
\includegraphics[width=52mm,height=15mm]{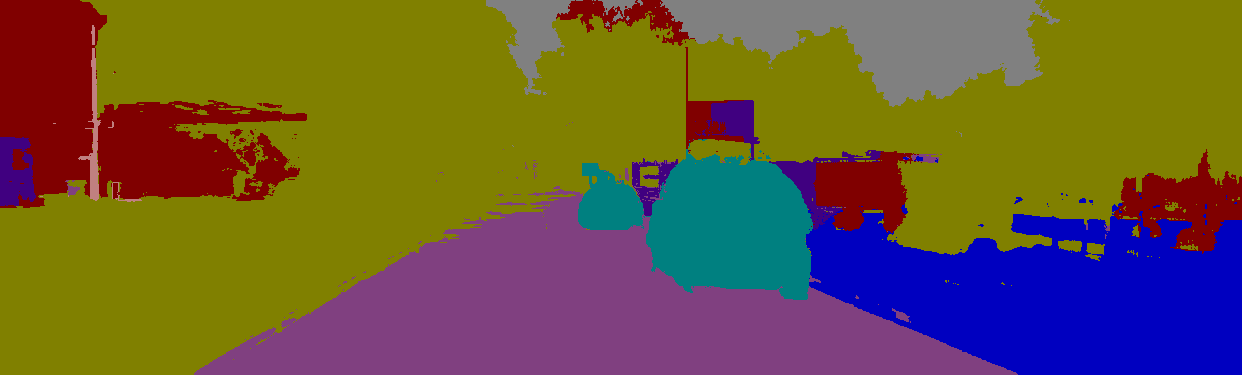} &
\includegraphics[width=52mm,height=15mm]{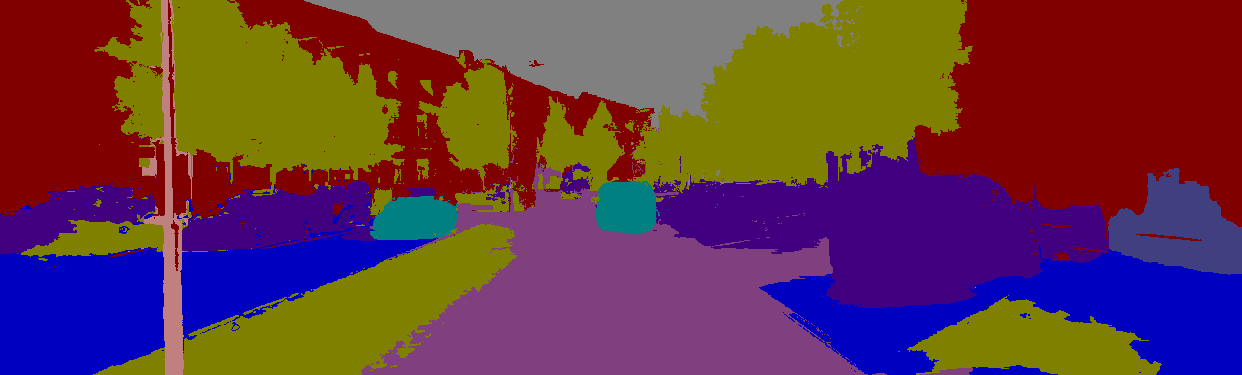} \\

\begin{sideways}\bf \centering   \ \  \ \ \ \ \   3D-REC \end{sideways} & 
\includegraphics[width=52mm,height=22mm]{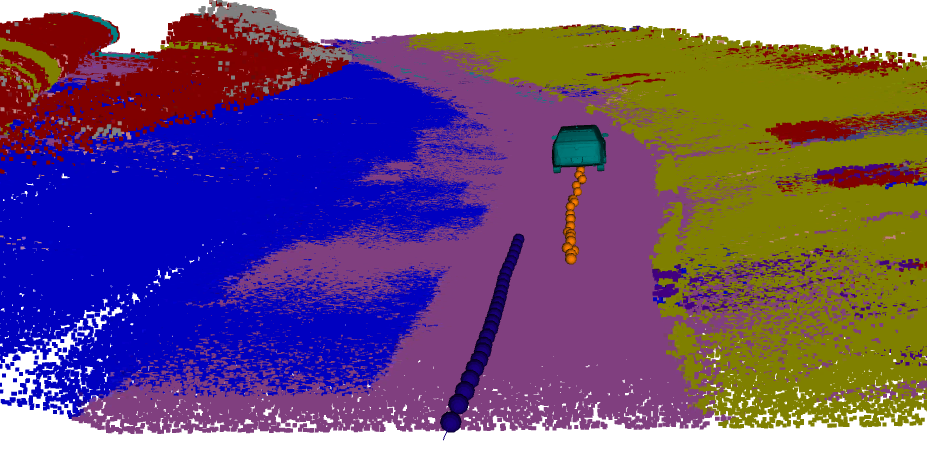} & 
\includegraphics[width=52mm,height=22mm]{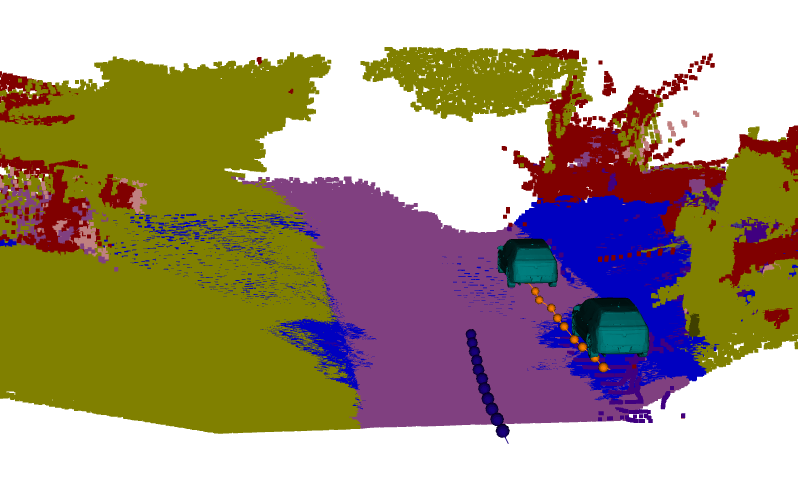} &
\includegraphics[width=52mm,height=22mm]{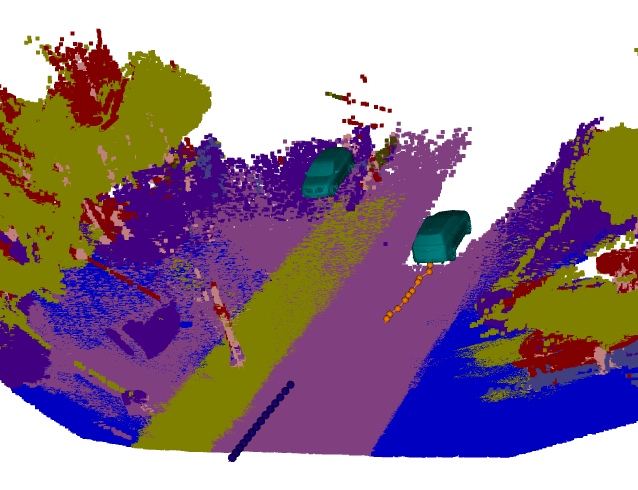}
\end{tabular}
\end{center}
 \caption{\small
We show the {\bf (INPUT)} image sequences for which we compute the semantic motion segmentation ({\bf SMS}). 
We have depicted the reconstruction of moving objects with their trajectories {\bf (3D-REC)}. Blue trajectories represent the camera capturing the scene. All segmentation color labels are consistent with
Figure~\ref{fig:pipeline1}. (best viewed in color)
}
\label{fig:figure_final}
\end{figure*}
 We compare the estimated trajectories of the moving objects to the extended Kalman filter based object tracking VISO2 (Stereo) ~\cite{lenz}. VISO2 S(Stereo) has reported error of 2.44 \% on the KITTI odometry dataset, making it a good baseline algorithm to compare with. 
 As proposed by Sturm et al. \cite{Sturm12iros}, the comparison methodology is based on ATE for root mean square error(RMSE), mean, median. We use their evaluation algorithm which aligns the 2 trajectories using SVD. We show the three statistics as mean and median are robust to outliers while RMSE shows the exact deviation from the ground truth.
 
The Table (\ref{tab:stationary_ate}) depicts the trajectory error estimation of the odometry of the KITTI1 sequence. the table shows the enhancement in the ATE error of the odometry with Motion segmentation and without motion segmentation. Similarly, Table (\ref{tab:KITTI2_ate}) depicts the trajectory error for moving object visible in all the 212 images of the sequence . This table depicts the improvement of the trajectory of moving objects with the help of semantic constraints imposed on the motion of the moving object.We show how each constraint on the motion of the moving object complement the trajectory computation and reconstruction of the dynamic objects.

For quantitative evaluation of our method, we have computed the trajectories of all the   moving objects. These trajectories are compared to their respective ground truth and the absolute position error of each pose is computed. We have done a histogram based evaluation of all the position error as depicted in Fig(\ref{fig:histogram}),here we compare the trajectories of our algorithm with VISO2. We have evaluated the algorithm for a complete of 297 poses of moving objects and found that our approach outperforms VISO2 and standard 2D bundle adjustment . The trajectories and reconstruction of some of the moving objects is depicted in the Fig(\ref{fig:figure_final}).   

\section{\bf CONCLUSION}
\label{sec:con}
In this paper, we have proposed a joint labelling framework for semantic motion segmentation and reconstruction 
in dynamic urban environments. We modelled the problem of creating a semantic dense map of moving objects in a 
urban environment using trajectory optimization . The experiments suggest that semantic segmentation  provide good 
initial estimates to aid generalized bundle adjustment based approach. This  helps in improving the localization 
of the moving objects and creates an accurate semantic map.

\begin{table}
\begin{tabular}{ p{2.5cm} | p{0.5cm} p{0.5cm} p{0.5cm} p{0.5cm} p{0.5cm} p{0.5cm}}
  Method & D & S & MS & SR & MR & SBA \\ \hline
  Sengupta et al.\cite{Sengupta_icra_2013} & \checkmark & \checkmark & & \checkmark & &\\
  Hane et all.\cite{conf/cvpr/HaneZCAP13} & \checkmark & \checkmark & & \checkmark & &\\
  Jianxiong et al.\cite{sun3d} & & & & \checkmark & &\\
  kundu et al.\cite{MulVSLAM_Abhijit_ICCV2011} & \checkmark & & \checkmark & & &\\
  valentin et al.\cite{conf/cvpr/ValentinSWST13} & \checkmark & \checkmark & & \checkmark &  &\\
  OURS & \checkmark & \checkmark & \checkmark & \checkmark & \checkmark & \checkmark\\
\end{tabular}
\caption{\small Comparision with related work. D=utdoor,S=Stereo camera ,MS=Motion segmentation,S=Semantic Reconstruction,MR=motion Reconstruction,SBA=Semantic Bundle adjustment}
\end{table}







\begin{thebibliography}{99}
\scriptsize
\bibitem{Sengupta_icra_2013} Sunando Sengupta and Eric Greveson and A. Shahrokni and Philip H.S. Torr, G. O. Young,Urban 3D Semantic Modelling Using Stereo Vision, in ICRA,2013.
\bibitem{Shotton06textonboost}  J. Shotton, J. Winn, C. Rother, and A. Criminisi,Textonboost: Joint appearance, shape and context modeling for multi-class object recognition and segmentation.In ECCV, pages 1-15, 2006.
\bibitem{conf/cvpr/HaneZCAP13} Hane, Christian and Zach, Christopher and Cohen, Andrea and Angst, Roland and Pollefeys, Marc, Joint 3D Scene Reconstruction and Class Segmentation. In CVPR ,2013, pages 97-104.
\bibitem{conf/cvpr/ValentinSWST13} Valentin, Julien P. C. and Sengupta, Sunando and Warrell, Jonathan and Shahrokni, Ali and Torr, Philip H. S., Mesh Based Semantic Modelling for Indoor and Outdoor Scenes.In CVPR 2013, pages 2067-2074.
\bibitem{sun3d} Jianxiong Xiao, Andrew Owens, Antonio Torralba, "SUN3D: A Database of Big Spaces Reconstructed Using SfM and Object Labels", in ICCV, 2013.
\bibitem{roussos2012dense} Roussos, Anastasios and Russell, Chris and Garg, Ravi and Agapito, Lourdes,Dense multibody motion estimation and reconstruction from a handheld camera, in ISMAR, 2012
\bibitem{YuanM06} Chang Yuan and Gerard Medioni,Reconstruction of Background and Objects Moving on Ground Plane Viewed from a Moving Camera, in CVPR, 2006.
\bibitem{MulVSLAM_Abhijit_ICCV2011} Kundu, Abhijit and Krishna, K. Madhava and Jawahar, C.V.,Realtime Multibody Visual SLAM with a Smoothly Moving Monocular Camera, in ICCV, 2011.
\bibitem{subspace2009} E.Elhamifar and R.Vidal. Sparse subspace clustering, in CVPR, 2009.


\bibitem{Sturm12iros} J. Sturm and N. Engelhard and F. Endres and W. Burgard and D. Cremers. A Benchmark for the Evaluation of RGB-D SLAM Systems,in IROS,2012.

\bibitem{LadickyRKT09}Ladicky Lubor,Russell Christopher,Kohli Pushmeet and Torr Philip H.S , Associative hierarchical CRFs for object class image segmentation, in ICCV,2009
\bibitem{Krahenbuhl_Koltun_2011}W. Choi and C. Pantofaru and S. Savarese,Efficient Inference in Fully Connected CRFs with Gaussian Edge Potentials,In NIPS,2011.
\bibitem{ceres}Sameer Agarwal and Keir Mierle,Ceres Solver: Tutorial \& Reference,Google

\bibitem{weinzaepfel:hal-00873592}Weinzaepfel, Philippe and Revaud, Jerome and Harchaoui, Zaid and Schmid, Cordelia,DeepFlow: Large displacement optical flow with deep matching,In ICCV,2013.

\bibitem{Romero-CanoN13}Victor Romero-Cano and Juan I. Nieto,Stereo-based motion detection and tracking from a moving platform,In IV,2013

\bibitem{lenz}Philip Lenz and Julius Ziegler and Andreas Geiger and Martin Roser,Sparse Scene Flow Segmentation for Moving Object Detection in Urban Environments,In IV,2011.

\bibitem{icvgippaper} N Dinesh Reddy, Prateek Singhal and K Madhava Krishna, Semantic Motion Segmentation Using Dense CRF Formulation, In ICVGIP,2014.

\bibitem{ZhouICCV2013Elastic}Qian-Yi Zhou and Stephen Miller and Vladlen Koltun,Elastic Fragments for Dense Scene Reconstruction,In ICCV,2013.
\bibitem{urtasun14eccv} K. Yamaguchi, D. McAllester and R. Urtasun, Efficient Joint Segmentation,  Occlusion Labeling, Stereo and Flow Estimation,In ECCV,2014.

 \bibitem{siddharth15icra} Siddharth Choudhary, Vadim Indelman, Henrik Christensen and Frank Dellaert, Information-based Reduced Landmark SLAM, In ICRA,2015.
 \bibitem{vedaldiVLFeat} A. Vedaldi and B. Fulkerson, VLFeat: An Open and Portable Library of
   Computer Vision Algorithms, 2008.

 \bibitem{matousek} {Matou\v{s}ek, Ji\v{r}\'{\i}}, On Variants of the Johnson-Lindenstrauss
   Lemma, Random Struct. Algorithms, 2008.

 \bibitem{dasgupta} {Anirban Dasgupta, Maxim Gurevich, Kunal Punera}, A Sparse Johnson-Lindenstrauss Transform, ACM Symposium on Theory of Computing, 2010.

 \bibitem{clarkson} {Kenneth Clarkson, David Woodruff}, Low Rank Approximation and Regression in Input Sparsity Time, ACM Symposium on Theory of Computing, 2012.

\bibitem{lepetit} {Vincen Lepetit, Moreno-Noguer Francesc and Pascal Fua}, EPnP: An Accurate O(n)   eolution to the PnP Problem, In IJCV, 2009.
\end{thebibliography}
\end{document}